\documentclass[10pt,twocolumn,letterpaper]{article}

\usepackage{cvpr}
\usepackage{times}
\usepackage{epsfig}
\usepackage{graphicx}
\usepackage{amsmath}
\usepackage{amssymb}

\usepackage{multirow}
\usepackage{bm}
\usepackage{xcolor}
\mathchardef\mhyphen="2D

\usepackage[pagebackref=true,breaklinks=true,letterpaper=true,colorlinks,bookmarks=false]{hyperref}

\cvprfinalcopy 

\def\cvprPaperID{3544} 

\ifcvprfinal\fi
\begin{document}

\title{Collaborative Spatiotemporal Feature Learning for Video Action Recognition}

\author{Chao Li \quad Qiaoyong Zhong \quad Di Xie \quad Shiliang Pu\\
Hikvision Research Institute\\
{\tt\small\{lichao15,zhongqiaoyong,xiedi,pushiliang\}@hikvision.com}
}

\maketitle

\begin{abstract}
  Spatiotemporal feature learning is of central importance for action recognition in videos. Existing deep neural network models either learn spatial and temporal features independently (C2D) or jointly with unconstrained parameters (C3D). In this paper, we propose a novel neural operation which encodes spatiotemporal features collaboratively by imposing a weight-sharing constraint on the learnable parameters. In particular, we perform 2D convolution along three orthogonal views of volumetric video data, which learns spatial appearance and temporal motion cues respectively. By sharing the convolution kernels of different views, spatial and temporal features are collaboratively learned and thus benefit from each other. The complementary features are subsequently fused by a weighted summation whose coefficients are learned end-to-end. Our approach achieves state-of-the-art performance on large-scale benchmarks and won the 1st place in the Moments in Time Challenge 2018. Moreover, based on the learned coefficients of different views, we are able to quantify the contributions of spatial and temporal features. This analysis sheds light on interpretability of the model and may also guide the future design of algorithm for video recognition.
\end{abstract}

\section{Introduction}
\label{sec:intro}

Recently, video action recognition has drawn increasing attention considering its potential in a wide range of applications such as video surveillance, human-computer interaction and social video recommendation. The key to this task lies in joint spatiotemporal feature learning. The spatial feature mainly describes appearance of objects involved in an action and the scene configuration as well within each frame of the video. Spatial feature learning is analogous to that of still image recognition, and thus easily benefits from the recent advancements brought by deep Convolutional Neural Networks (CNN)~\cite{krizhevsky2012imagenet}. While the temporal feature captures motion cues embedded in the evolving frames over time. There are two challenges that arise. One is how to learn the temporal feature. The other is how to properly fuse spatial and temporal features.

The first attempt of researchers is to model temporal motion information explicitly and in parallel to spatial information. Raw frames and optical flow between adjacent frames are exploited as two input streams of a deep neural network~\cite{simonyan2014two-stream,Feichtenhofer2017Spatiotemporal}. On the other hand, as a generalization of 2D ConvNets (C2D) for still image recognition, 3D ConvNets (C3D) are proposed to tackle 3D volumetric video data~\cite{Du2014Learning}. In C3D, spatial and temporal features are closely entangled and jointly learned. That is, rather than learning spatial and temporal features separately and fusing them at the top of the network, joint spatiotemporal features are learned by 3D convolutions distributed over the whole network. Considering the excellent feature representation learning capability of CNN, ideally C3D should achieve great success on video understanding just as C2D does on image recognition. However, the huge number of model parameters and computational inefficiency limit the effectiveness and practicality of C3D.

\begin{figure}[t]
  \begin{center}
  \includegraphics[width=0.9\linewidth]{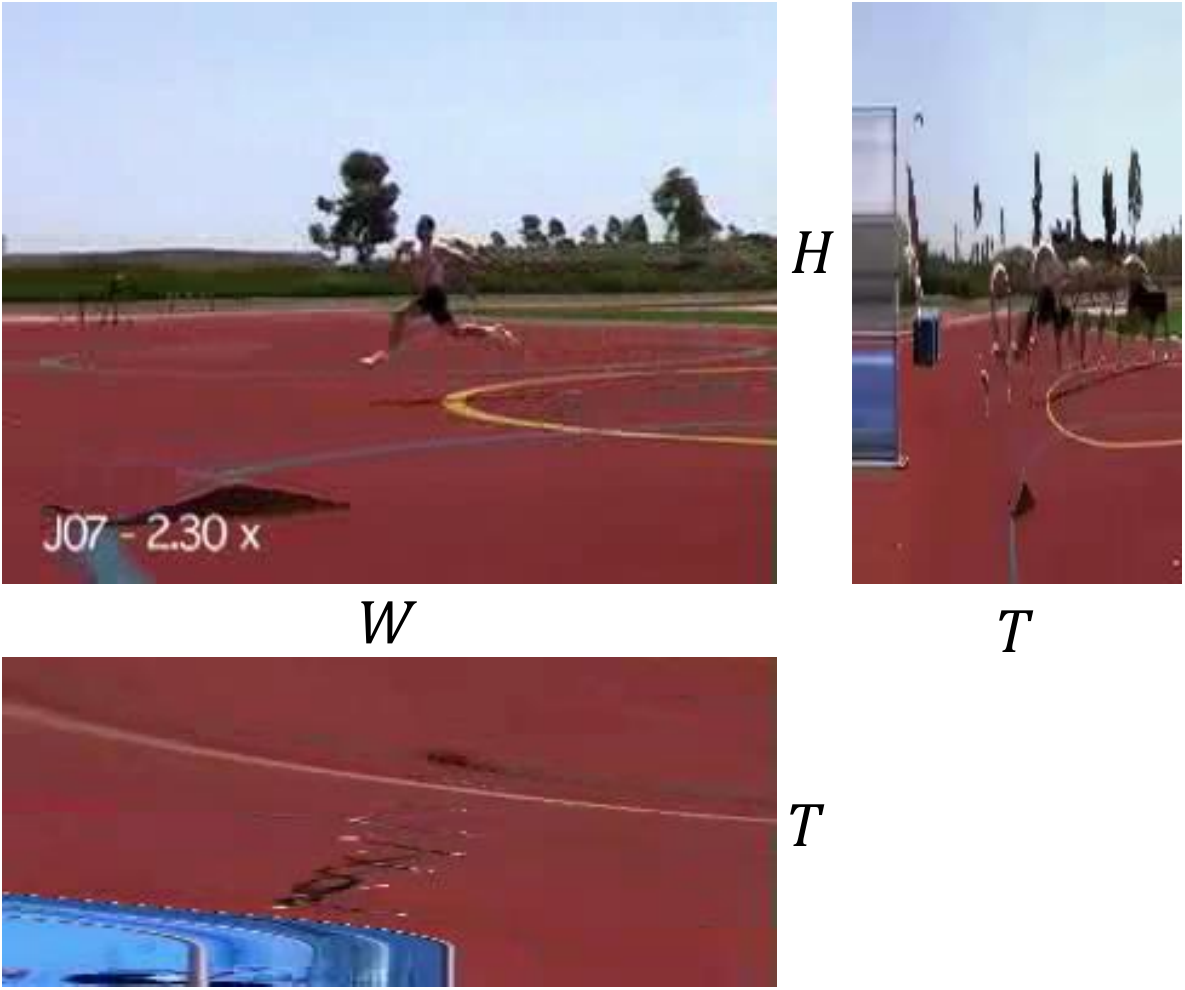}
  \end{center}
  \caption{Visualization of three views of a video, which motivates our design of collaborative spatiotemporal feature learning. Top left: view of $H\mhyphen W$. Top right: view of $T\mhyphen H$. Bottom: view of $T\mhyphen W$.
  \label{fig:conception}}
\end{figure}

\begin{figure}[t]
  \begin{center}
  \includegraphics[width=\linewidth]{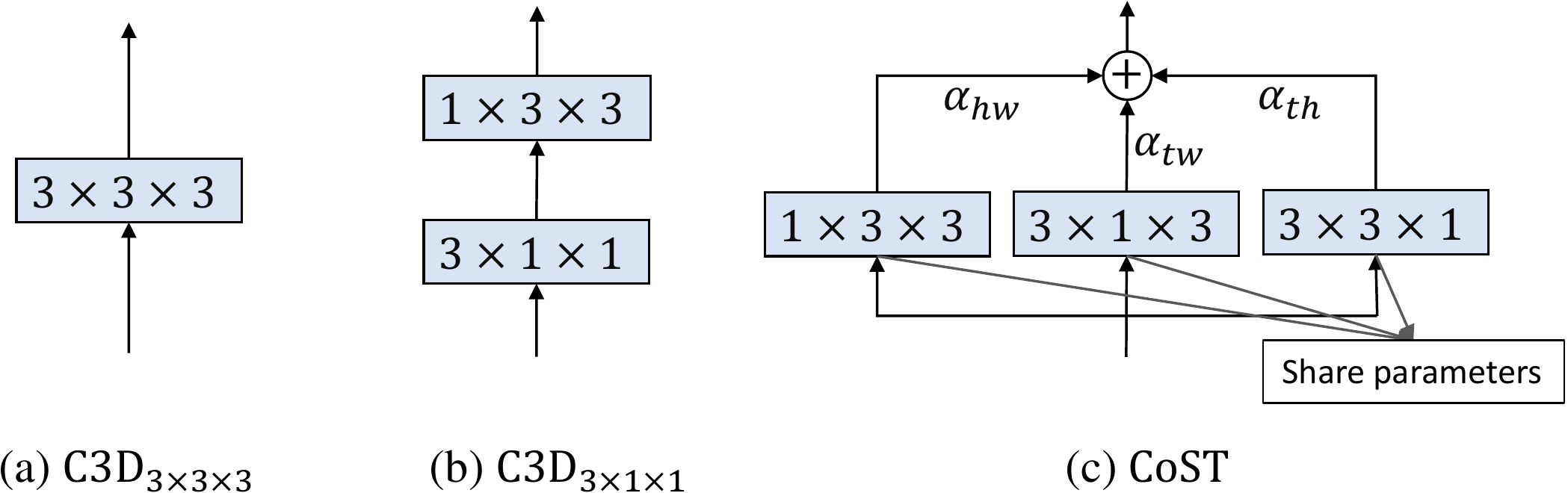}
  \end{center}
  \caption{Comparison of CoST to common spatiotemporal feature learning architectures. (a) C3D$_{3\times 3\times 3}$. (b) C3D$_{3\times 1\times 1}$. (c) The proposed CoST.
  \label{fig:networks}}
\end{figure}

In this paper, we propose a novel Collaborative SpatioTemporal (CoST) feature learning operation, which learns spatiotemporal features jointly with a weight-sharing constraint. Given a 3D volumetric video tensor, we flatten it into three sets of 2D images by viewing it from different angles. Then 2D convolution is applied to each set of 2D images. Figure~\ref{fig:conception} shows the 2D snapshots from three views of an exemplary video clip, where a man is high jumping at the stadium. View of $H\mhyphen W$ is the natural view with which human beings are familiar. By scanning the video frame by frame from this view over time $T$, we are able to understand the video content. Although snapshots from views involving $T$ (i.e. $T\mhyphen W$ and $T\mhyphen H$) are difficult to interpret for human beings, they contain exactly the same amount of information as the normal $H \mhyphen W$ view. More importantly, rich motion information is embedded \emph{within each frame} rather than between frames. Hence 2D convolutions on frames of the $T\mhyphen W$ and $T\mhyphen H$ views are able to capture temporal motion cues directly. As shown in Figure~\ref{fig:networks}(c), by fusing complementary spatial and temporal features of the three views, we are able to learn spatiotemporal features using 2D convolutions rather than 3D convolutions.

Notably, the convolution kernels of different views are shared for the following reasons. 1) From the visualization of the frames of different views (see Figure~\ref{fig:conception}), their visual appearances are compatible. For example, common spatial patterns such as edges and color blobs also exist in temporal views ($T\mhyphen H$ and $T\mhyphen W$). Hence, the same set of convolution kernels can be applied on frames of different views. 2) Convolution kernels in C2D networks are inherently redundant without pruning~\cite{He_2017_ICCV,li2017pruning,xie2017all}. While the redundant kernels can be exploited for temporal feature learning by means of weight sharing. 3) The number of model parameters is greatly reduced, such that the network is easier to train and less prone to overfitting, resulting in better performance. Besides, the success of spatial feature learning on still images (\eg carefully designed network architecture and pre-trained parameters) can be transferred to temporal domain with little effort.

The complementary features of different views are fused by a weighted summation. We learn an independent coefficient for each channel in each view, which allows the network to attend to either spatial or temporal features on demand. Moreover, based on the learned coefficients, we are able to quantify the respective contributions of spatial domain and temporal domain.

Based on the CoST operation, we build a convolutional neural network. We will henceforth refer to both the operation and the network as CoST, which should be easy to identify according to its context. Compared with C2D, CoST can learn spatiotemporal features jointly. While compared with C3D, CoST is based on 2D rather than 3D convolutions. CoST essentially bridges the gap between C2D and C3D, where the benefits from both sides, i.e. compactness of C2D and representation capability of C3D are retained. For the task of action recognition in videos, experiments show that CoST achieves superior performance over both C2D and C3D.

The main contributions of this work are summarized as follows:
\begin{itemize}
  \item We propose CoST, which collaboratively learns spatiotemporal features using 2D convolutions rather than 3D convolutions.
  \item To the best of our knowledge, this is the first work on quantitative analysis of importance of spatial and temporal features for video understanding.
  \item The proposed CoST model outperforms the conventional C3D model and its variants, achieving state-of-the-art performance on large-scale benchmarks.
\end{itemize}

\section{Related Work}
In the early stage, hand-crafted representations have been well explored for video action recognition. Many feature descriptors for 2D images are generalized to 3D spatiotemporal domain, \eg Space-Time Interest Points (STIP)~\cite{Laptev2005On}, SIFT-3D~\cite{Scovanner2007A}, Spatiotemporal SIFT~\cite{ST-SIFT} and 3D Histogram of Gradient~\cite{Klaser2010A}. The most successful hand-crafted representations are dense trajectories~\cite{Wang2013Dense} and its improved version~\cite{Wang2014Action}, which extract local features along trajectories guided by optical flow.

Encouraged by the great success of deep learning, especially the CNN model for image understanding, there are a number of attempts to develop deep learning methods for action classification~\cite{zhao2017temporal}. The two-stream architecture~\cite{simonyan2014two-stream} utilizes visual frames and optical flows between adjacent frames as two separate inputs of the network, and fuses their output classification scores as the final prediction. Many works follow and extend this architecture~\cite{feichtenhofer2016spatiotemporal,Feichtenhofer2017Spatiotemporal,zhu2017hidden}. The LSTM networks have also been employed to capture temporal dynamics and long range dependences in videos. In~\cite{ng2015beyond,donahue2017long-term} CNN is used to learn spatial feature for each frame, while LSTM is used to model temporal evolutions.

More recently, with the increasing computing capability of modern GPUs and the availability of large-scale video datasets, 3D ConvNet (C3D) has drawn more and more attention. In~\cite{Du2014Learning} a 11-layer C3D model is designed to jointly learn spatiotemporal features on the Sports-1M dataset~\cite{sports-1M}. However, the huge computational cost and the dense parameters of C3D make it infeasible to train a very deep model. Qiu \etal~\cite{Qiu2017Learning} proposed Pseudo-3D (P3D) which decomposes a 3D convolution of $3\times 3\times 3$ into a 2D convolution of $1\times 3\times 3$ followed by a 1D convolution of $3\times 1\times 1$. In another work~\cite{Tran2017A}, similar architecture is explored and referred to as (2+1)D. \cite{kinetics} proposed the Inflated 3D ConvNet (I3D), which is exactly C3D whose parameters are initialized by inflating the parameters of pre-trained C2D model.

The most closely related work to ours is Slicing CNN~\cite{Shao_2016_CVPR}, which also learns features from multiple views for crowd video understanding. However, there are substantial differences between Slicing CNN and the proposed CoST. Slicing CNN learns independent features of the three views via three different network branches, which are merged at the top of the network. Aggregation of spatial and temporal features is conducted only once at the network level. On the contrary, we learn spatiotemporal features collaboratively using a novel CoST operation. Spatiotemporal feature aggregation is conducted layer-wise.

\section{Method}
In this section, we first review the conventional C2D and C3D architectures, which are implemented as a baseline. Then we introduce the proposed CoST. The connection and comparison between CoST and C2D / C3D are also discussed.

\begin{figure}[t]
  \begin{center}
  \includegraphics[width=0.9\linewidth]{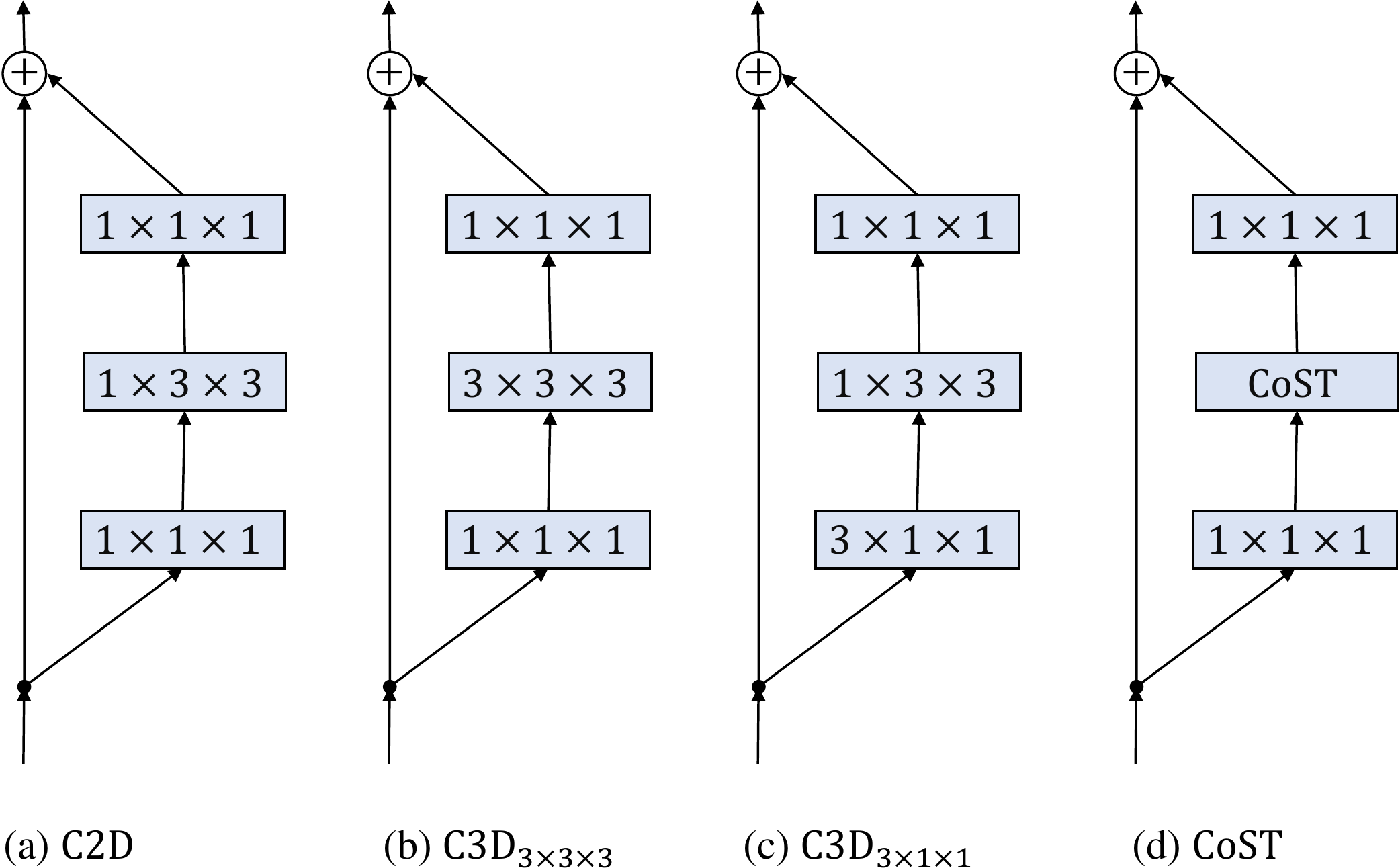}
  \end{center}
  \caption{Comparison of various residual units for action recognition in videos.
  \label{fig:res_blocks}}
\end{figure}

\subsection{2D ConvNets}
\newcommand{\resblock}[3]{
  \multirow{3}{*}{
    $\begin{bmatrix}
      1\times1\times1,#1\\1\times3\times3,#1\\1\times1\times1,#2
    \end{bmatrix}\times#3$
  }
}
\begin{table}[t]
  \small
  \begin{center}
  \begin{tabular}[b]{c|c|c|c}
    \hline
    Name& Output Size& Filter&Stride\\ \hline
    input& 8$\times$224$\times$224& none &none \\ \hline
    conv$_1$& 8$\times$112$\times$112& $1\times 7\times 7,64$ &1,2,2 \\ \hline
    pool$_1$& 8$\times$56$\times$56& $3\times 3\times 3,max$ &1,2,2 \\ \hline
    \multirow{3}{*}{block$_1$}  & \multirow{3}{*}{8$\times$56$\times$56}& \resblock{64}{256}{3} &\multirow{3}{*}{1,1,1}\\ 
    & & & \\
    & & & \\ \hline
    pool$_2$& 4$\times$56$\times$56& $3\times 1\times 1,max$ &2,1,1 \\ \hline
    \multirow{3}{*}{block$_2$}  & \multirow{3}{*}{4$\times$28$\times$28}& \resblock{128}{512}{4} &\multirow{3}{*}{1,2,2}\\ 
    & & &\\
    & & &\\\hline
    \multirow{3}{*}{block$_3$}  & \multirow{3}{*}{4$\times$14$\times$14}& \resblock{256}{1024}{6} &\multirow{3}{*}{1,2,2}\\
    & & & \\
    & & & \\\hline
    \multirow{3}{*}{block$_4$}  & \multirow{3}{*}{4$\times$7$\times$7}& \resblock{512}{2048}{3} &\multirow{3}{*}{1,2,2}\\
    & & & \\
    & & & \\\hline
    pool$_3$& 1$\times$1$\times$1& $4\times 7\times 7,average$ &1,1,1 \\ \hline
    fc& 1$\times$1$\times$1& 2048$\times$class &1,1,1 \\ \hline
  \end{tabular}
  \end{center}
  \caption{Architecture of ResNet-50-C2D. Spatial striding is performed on the first residual unit of each block.}
  \label{table:resnet50-c2d}
\end{table}
C2D leverages the strong spatial feature representation capability of 2D convolutions, while simple strategy (\eg pooling) is utilized for temporal feature aggregation. In this work, we implement C2D as a baseline model. We choose ResNets~\cite{He2015Deep} as our backbone networks, whose residual unit is shown in Figure~\ref{fig:res_blocks}(a). To handle 3D volumetric video data, the vanilla ResNets need to be adapted accordingly. Taking ResNet-50 as an example, its adapted version for video action recognition is illustrated in Table~\ref{table:resnet50-c2d}. For convenience we will henceforth refer to it as ResNet-50-C2D. Note the differences between ResNet-50-C2D and vanilla ResNet-50. Firstly, all $k\times k$ 2D convolutions are adapted to their 3D form, i.e. $1\times k \times k$. Secondly, a temporal pooling (pool$_2$) is append after block$_1$ to halve the number of frames from 8 to 4. Thirdly, the global average pooling (pool$_3$) is also adapted from $7\times 7$ to $4\times 7\times 7$ such that spatial and temporal features are aggregated simultaneously. Similarly, we can setup ResNet-101-C2D based on ResNet-101.

\subsection{3D ConvNets}
C3D is a natural generalization of C2D for 3D video data. In C3D, 2D convolutions are converted to 3D by inflating the filters from square to cubic. For example, an $h\times w$ 2D filter can be converted into a $t\times h\times w$ 3D filter by introducing an additional temporal dimension $t$~\cite{feichtenhofer2016spatiotemporal,kinetics}. In modern deep CNN architectures like ResNets, there are two main types of filters, i.e. $1\times 1$ and $3\times 3$. As explored in~\cite{Wang2017Non}, given a residual unit comprised of $1\times 1$ and $3\times 3$ convolutions, we may either inflate the middle $3\times 3$ filter into $3\times 3\times 3$ (C3D$_{3\times 3\times 3}$) as shown in Figure~\ref{fig:res_blocks}(b), or inflate the first $1\times 1$ filter into $3\times 1\times 1$ (C3D$_{3\times 1\times 1}$) as shown in Figure~\ref{fig:res_blocks}(c). Experiments in~\cite{Wang2017Non} demonstrate that C3D$_{3\times 3\times 3}$ and C3D$_{3\times 1\times 1}$ achieve comparable performance, while the latter contains much fewer parameters and is more computationally efficient. Therefore, in our implementation, C3D$_{3\times 1\times 1}$ is adopted and referred to as C3D for simplicity. Notably, the C3D$_{3\times 1\times 1}$ model learns spatial and temporal features alternatively rather than jointly, which is very similar to the (2+1)D~\cite{Tran2017A} and P3D~\cite{Qiu2017Learning} models.

In our implementation, we inflate the first $1\times 1$ filter for every two residual units following~\cite{Wang2017Non}. However, we leave conv$_1$ unchanged to be 2D ($1\times 7\times 7$), as opposed to~\cite{Wang2017Non}.

\subsection{CoST}

In this section, we elaborately describe the proposed CoST model. Figure~\ref{fig:networks} compares the proposed CoST operation to common spatiotemporal feature aggregating modules. As mentioned above, C3D$_{3\times 3\times 3}$ utilizes a 3D convolution of $3\times 3\times 3$ to extract spatial (along $H$ and $W$) and temporal (along $T$) features jointly. In the C3D$_{3\times 1\times 1}$ configuration, a 1D $3\times 1\times 1$ convolution along $T$ is utilized to aggregate temporal feature, followed by a 2D $1\times 3\times 3$ convolution along $H$ and $W$ for spatial feature. While in the proposed method, we perform 2D $3\times 3$ convolutions along three views of the $T\times H\times W$ volumetric data, i.e. $H\mhyphen W$, $T\mhyphen H$ and $T\mhyphen W$ separately. Notably, the parameters of the three-view convolutions are shared, which keeps the number of parameters the same as single-view 2D convolution. The three resulting feature maps are subsequently aggregated with weighted summation. The weights are also learned during training in an end-to-end manner.

Let $\bm{x}$ denote the input feature maps of size $T\times H\times W\times C_1$ where $C_1$ is the number of input channels. The three sets of output feature maps from different views are computed by:
\begin{equation}\label{eq:conv_view}
  \begin{split}
    \bm{x}_{hw}&=\bm{x}\otimes \bm{w}_{1\times3\times3}, \\
    \bm{x}_{tw}&=\bm{x}\otimes \bm{w}_{3\times1\times3}, \\
    \bm{x}_{th}&=\bm{x}\otimes \bm{w}_{3\times3\times1}, 
  \end{split}
\end{equation}
where $\otimes$ denotes 3D convolution, $\bm{w}$ is convolution filters of size $3\times 3$ shared among the three views. To apply $\bm{w}$ to frames of different views, we insert an additional dimension of size $1$ at different indices. The resulting variants of $\bm{w}$, i.e. $\bm{w}_{1\times3\times3}$, $\bm{w}_{3\times1\times3}$ and $\bm{w}_{3\times3\times1}$ learn features of the $H\mhyphen W$, $T\mhyphen W$ and $T\mhyphen H$ views respectively. Then, the three sets of feature maps are aggregated with weighted summation:
\begin{equation}\label{eq:sum_views}
  y=
  \begin{bmatrix}\alpha_{hw},\alpha_{tw},\alpha_{th}
  \end{bmatrix}
  \begin{bmatrix}\bm{x}_{hw}\\ \bm{x}_{tw}\\\bm{x}_{th}
  \end{bmatrix},
\end{equation}
where $\bm{\alpha}=[\alpha_{hw},\alpha_{tw},\alpha_{th}]$ are the coefficients of size $C_2\times 3$. $C_2$ is the number of output channels and $3$ denotes three views. To avoid magnitude explosion of the resulting responses from multiple views, $\bm{\alpha}$ is normalized with the Softmax function along each row.

\begin{figure}[t]
  \begin{center}
  \includegraphics[width=0.8\linewidth]{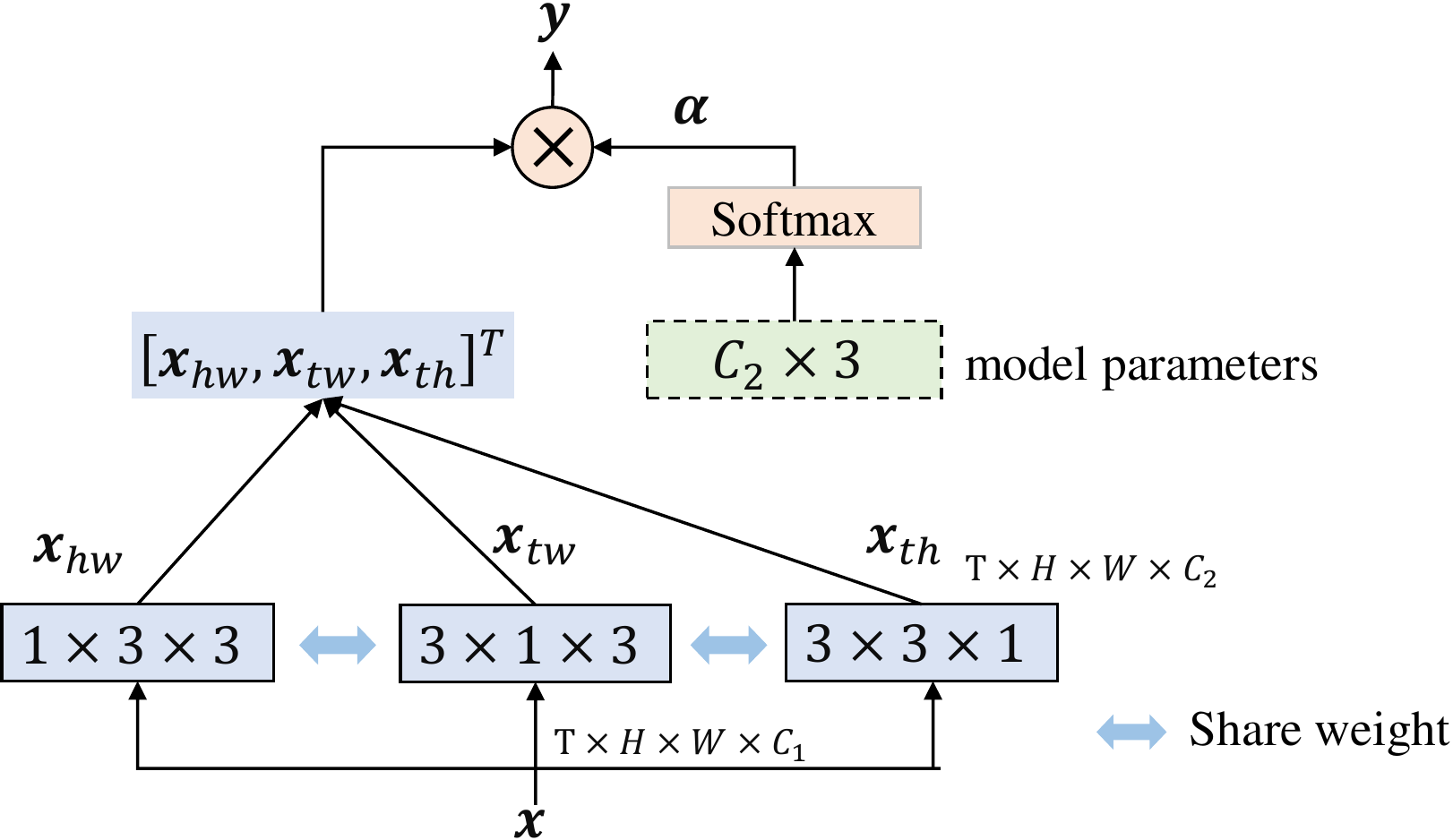}
  \end{center}
  \caption{Architecture of CoST(a), where the coefficients $\bm{\alpha}$ are part of the model parameters.
  \label{fig:mv_a}}
\end{figure}

\begin{figure}[t]
  \begin{center}
  \includegraphics[width=0.9\linewidth]{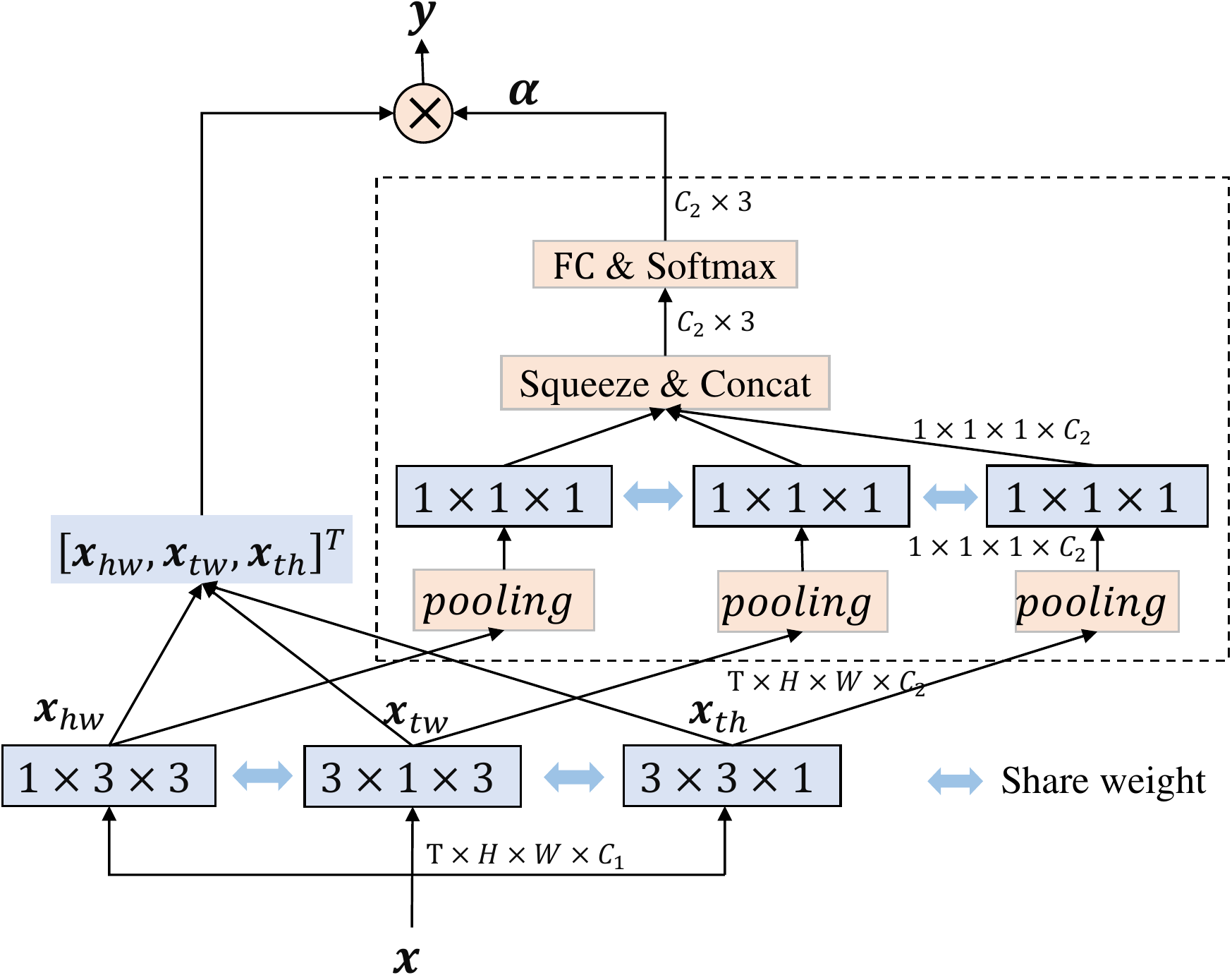}
  \end{center}
  \caption{Architecture of CoST(b), where the coefficients $\bm{\alpha}$ are predicted by the network.
  \label{fig:mv_b}}
\end{figure}

To learn the coefficients $\bm\alpha$, we propose two architectures, named CoST(a) and CoST(b).

\noindent\textbf{CoST(a).} As illustrated in Figure~\ref{fig:mv_a}, the coefficients $\bm{\alpha}$ are considered as part of the model parameters, which can be updated with back-propagation during training. During inference, the coefficients are fixed and the same set of coefficients is applied to each video clip.

\noindent\textbf{CoST(b).} The coefficients $\bm{\alpha}$ are predicted by the network based on the feature maps by which $\bm\alpha$ will be multiplied. This design is inspired by the recent \emph{self-attention}~\cite{vaswani2017attention} mechanism for machine translation. In this case, the coefficients for each sample depend on the sample itself. It can be formulated as:
\begin{equation}\label{eq:predict_alpha}
    \begin{bmatrix}\alpha_{hw},\alpha_{tw},\alpha_{th}
    \end{bmatrix}
    =f(
    \begin{bmatrix}\bm{x}_{hw},\bm{x}_{tw},\bm{x}_{th}
    \end{bmatrix})
\end{equation}
The architecture of CoST(b) is illustrated in Figure~\ref{fig:mv_b}. The computational block inside the dashed lines represents the function $f$ in Equation~\eqref{eq:predict_alpha}. Specifically, for each view, we first reduce the feature map from a size of $T\times H\times W\times C_2$ to $1\times 1\times 1\times C_2$ using global max pooling along dimension $T$, $H$ and $W$. Then, a $1\times 1\times 1$ convolution is applied on the pooled features, whose weights are also shared by all three views. This convolution maps features of dimension $C_2$ back to $C_2$, which captures the contextual information among channels. After that, the three sets of features are concatenated and fed into a fully connected (FC) layer. As opposed to the $1\times 1\times 1$ convolution, this FC layer is applied to each row of the $C_2 \times 3$ matrix, which captures the contextual information among different views. Finally, we normalize the output by the Softmax function.
  
The residual unit of the proposed CoST is shown in Figure~\ref{fig:res_blocks}(d). We replace the middle $3\times 3$ convolution with our CoST operation, either CoST(a) or CoST(b), and leave the preceding $1\times 1$ convolution unchanged. Based on the C2D configuration of ResNets, we build CoST by replacing the C2D unit with the proposed CoST unit for every two residual units, which is consistent to C3D.

\subsection{Connection to C2D and C3D}
\label{sec:connection}
\begin{figure}[t]
  \begin{center}
  \includegraphics[width=\linewidth]{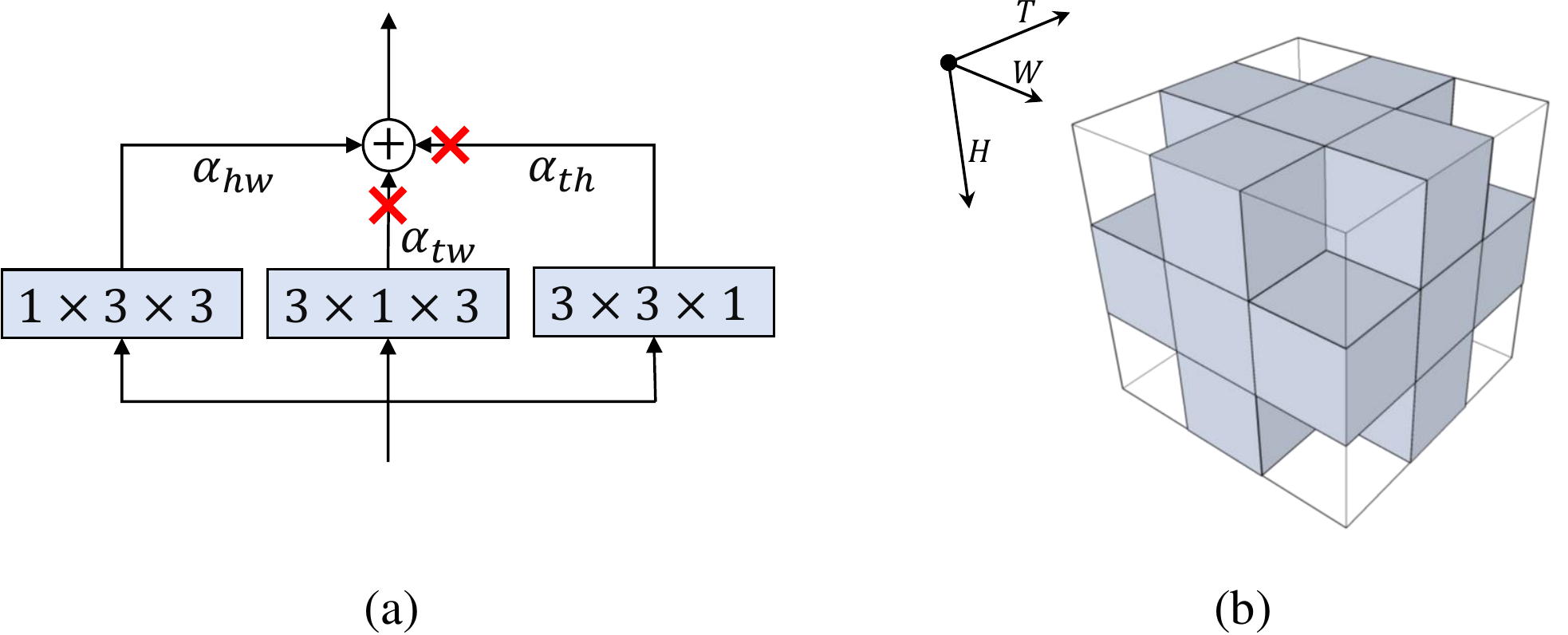}
  \end{center}
  \caption{Connection of CoST to C2D (a) and C3D (b). 
  \label{fig:connection}}
\end{figure}
The proposed CoST is closely related to C2D and C3D. As shown in Figure~\ref{fig:connection}(a), if the coefficients of the $T\mhyphen W$ and $T\mhyphen W$ views were set to zero, CoST degenerates to C2D. Hence, CoST is a strict generalization of C2D. 

To compare CoST with C3D, let us exclude the dimensions of input and output channels for simplicity. 3D convolution with a kernel size of $k\times k \times k$ contains $k^3$ parameters and covers a cubic receptive field of $k^3$ voxels. While the proposed CoST operation covers an irregular receptive field of $3k^2-3k+1$ voxels. Figure~\ref{fig:connection}(b) shows a comparison of receptive field when $k$ is equal to 3. C3D covers the whole $3\times 3\times 3$ cube, while CoST covers the shaded region excluding the 8 corner voxels. If the convolution kernels of the three views are learned separately without weight sharing, CoST is nearly equivalent to C3D except that the 8 corner parameters of the cubic kernel are fixed to zero and not learnable. When weight sharing is enabled in CoST, although the receptive field contains 19 voxels in total, the corresponding 19 parameters can be derived from the 9 learnable parameters shared among different views. Therefore, CoST can be considered as a special case of C3D, where similar receptive field is covered with significantly reduced number of parameters.

In terms of computational cost, CoST is also superior over C3D. The number of multiply-adds involved in the CoST operation is approximately $3k^2$ (excluding input and output channels), while that of C3D is $k^3$. Computational cost of CoST increases quadratically with the kernel size rather than cubically. This characteristic makes the employment of large kernel possible, which has not been explored yet on video data. Moreover, for the CoST(a) variant, some voxels in the receptive field are duplicately computed by multiple views in our current implementation. With an optimized implementation, the number of multiply-adds can be reduced from $3k^2$ to $3k^2-3k+1$, \eg from 27 to 19 (save $\sim 30\%$) for the case of $k=3$.

\section{Experiments}
To validate the effectiveness of the proposed CoST for the task of action recognition in videos, we perform extensively experiments on two of the largest benchmark datasets, i.e. Moments in Time~\cite{monfortmoments} and Kinetics~\cite{kinetics}. Accuracies are measured on the validation set of both datasets in all experiments.

\subsection{Datasets}
\noindent\textbf{Moments in Time.}
The Moments in Time dataset contains 802245 training videos and 39900 validation videos from 339 action categories. The videos are trimmed such that the duration is about 3 seconds.

\noindent\textbf{Kinetics.}
The Kinetics dataset contains 236763 training videos and 19095 validation videos, which are annotated as one of 400 human action categories. Note that the full Kinetics dataset contains a bit more samples. The numbers only cover the samples we are able to download. The duration of the videos is about 10 seconds.

\subsection{Implementation Details}
During training, we first sample 64 continuous frames from a video and then sub-sample one frame for every 8 frames, resulting in 8 frames in total. Next, image patches with a size of $224\times 224$ pixels are randomly cropped from a scaled video whose shorter side is randomly sampled between 256 and 320 pixels. Hence, the network input is of dimension $8\times 224 \times 224$. In all experiments, our models are initialized from ImageNet~\cite{russakovsky2015imagenet} pre-trained 2D models. We train the models on an 8-GPU machine. To speedup training, the 8 GPUs are grouped into two workers and the weights are updated asynchronously between the two workers. Each GPU process a mini-batch of 8 video clips. That is, for each worker 4 GPUs are employed, resulting in a total mini-batch size of 32. We train the models for 600k iterations using the SGD optimizer with momentum. We use a momentum of 0.9 and a weight decay of 0.0001. The learning rate is initialized to 0.005 and reduced by a factor of 10 at 300k and 450k iterations respectively.

During inference, following~\cite{Wang2017Non} we perform spatially fully convolutional inference on videos whose shorter side is rescaled to 256 pixels. While for the temporal domain, we sample 10 clips evenly from a full-length video and compute their classification scores individually. The final prediction is the averaged score of all clips.

\subsection{Ablation Studies}
To validate the effectiveness of individual components of our approach, we perform ablation studies on coefficient learning, impact of collaborative spatiotemporal feature learning and improvements of CoST over C2D and C3D.

\subsubsection{Coefficient Learning}
\begin{table}[t]
  \begin{center}
  \begin{tabular}[b]{lc|ccc}
    \hline
    \multirow{2}{*}{Dataset} & \multirow{2}{*}{Method} & \multicolumn{3}{c}{Accuracy (\%)} \\
     &  &Top-1 &Top-5 &Average\\ \hline
    \multirow{2}{*}{Moments} & CoST(a) &29.3 &55.8 &42.6\\
    & CoST(b) &\textbf{30.1}  &\textbf{57.2} &\textbf{43.7}\\ \hline
    \multirow{2}{*}{Kinetics} & CoST(a) &73.6 &90.8 &82.2\\
    & CoST(b) &\textbf{74.1}  &\textbf{91.2} &\textbf{82.7}\\ \hline
  \end{tabular}
  \end{center}
  \caption{Comparison of CoST(a) and CoST(b) for coefficient learning. The backbone network is ResNet-50.}
  \label{table:mv_a_b}
\end{table}

We first compare the performance of the two CoST variants for coefficient learning of different views. As shown in Table~\ref{table:mv_a_b}, on both of the Moments in Time and Kinetics datasets, coefficients predicted by the network (CoST(b)) outperform those learned as model parameters (CoST(a)). This result verifies the effectiveness of the self-attention mechanism introduced in our model. It also reveals that for different video clips, the importance of spatial and temporal features varies. Henceforth, the CoST(b) architecture is adopted in the following experiments.

\subsubsection{Impact of Collaborative Feature Learning}
To validate the effectiveness of collaborative spatiotemporal feature learning through weight sharing, we compare the results of the CoST(b) network with and without weight sharing. When weight sharing is disabled, the parameters of the three convolutional layers in Figure~\ref{fig:mv_b} are learned independently such that spatiotemporal features are learned in a decoupled manner. As listed in Table~\ref{table:weight-share}, with weight sharing among different views, accuracies get improved by about 1\% on both datasets. This result shows that our analysis on the characteristics of the three spatial and temporal views in Section~\ref{sec:intro} is reasonable and their collaborative feature learning is beneficial.

\begin{table}[t]
  \begin{center}
  \begin{tabular}[b]{lc|ccc}
    \hline
    \multirow{2}{*}{Dataset} & \multirow{2}{*}{Share Weight} & \multicolumn{3}{c}{Accuracy (\%)} \\
     &  & Top-1 & Top-5 &Average\\ \hline
    \multirow{2}{*}{Moments} &  &29.0  &56.1 &42.5\\ 
    & \checkmark &\textbf{30.1} &\textbf{57.2} &\textbf{43.7}\\\hline
    \multirow{2}{*}{Kinetics} &  &73.2 &90.2 &81.7\\
    & \checkmark &\textbf{74.1} &\textbf{91.2} &\textbf{82.7}\\ \hline
  \end{tabular}
  \end{center}
  \caption{Performance improvements brought by weight sharing using ResNet-50 as the backbone.}
  \label{table:weight-share}
\end{table}

\subsubsection{Improvements over C2D and C3D}
To compare CoST with the C2D and C3D baselines, we train all the three networks using the same protocol. Their performances on the Moments in Time and Kinetics datasets are listed in Table~\ref{table:moments_result} and Table~\ref{table:kintics_result} respectively. We can see that C3D is far better than C2D, while CoST consistently outperforms C3D by about 1\%, which clearly demonstrates the superiority of CoST. Note that the performance of C3D with ResNet-50 backbone is on par with the proposed CoST without weight sharing (see Table~\ref{table:weight-share}), which validates the connection between CoST and C3D described in Section~\ref{sec:connection}.

\begin{table}[t]
  \begin{center}
  \begin{tabular}[b]{ll|ccc}
    \hline
    \multicolumn{2}{c|}{\multirow{2}{*}{Method}} & \multicolumn{3}{c}{Accuracy (\%)} \\
    &  & Top-1 & Top-5 &Average\\ \hline
    \multirow{3}{*}{ResNet-50} &C2D & 27.9 &54.6 &41.3\\
    & C3D & 29.0 &55.3 & 42.2\\
    & CoST  & \textbf{30.1} &\textbf{57.2} & \textbf{43.7}\\ \hline
    \multirow{3}{*}{ResNet-101} & C2D &30.0 &56.8 &43.4\\
    & C3D &30.6 &57.7 &44.2\\
    & CoST  &\textbf{31.5} &\textbf{57.9} &\textbf{44.7}\\
    \hline
  \end{tabular}
  \end{center}
  \caption{Performance comparison of C2D, C3D and CoST on the validation set of Moments in Time.}
  \label{table:moments_result}
\end{table}

\begin{table}[t]
  \begin{center}
  \begin{tabular}[b]{ll|ccc}
    \hline
    \multicolumn{2}{c|}{\multirow{2}{*}{Method}} & \multicolumn{3}{c}{Accuracy (\%)} \\
    &  & Top-1 & Top-5 &Average\\ \hline
    \multirow{3}{*}{ResNet-50} &C2D & 71.5 &89.8 &80.7\\
    &C3D & 73.3 &90.4 &81.9\\
    &CoST  & \textbf{74.1} &\textbf{91.2} &\textbf{82.7}\\ \hline
    \multirow{3}{*}{ResNet-101} & C2D &72.9 &89.8 &81.4\\
    &C3D &74.5 &91.1 &82.8\\
    &CoST  &\textbf{75.5} &\textbf{92.0} &\textbf{83.8}\\
    \hline
  \end{tabular}
  \end{center}
  \caption{Performance comparison of C2D, C3D and CoST on the validation set of Kinetics.}
  \label{table:kintics_result}
\end{table}

\subsection{Comparisons with the State-of-the-arts}
\begin{table*}[t]
  \begin{center}
  \begin{tabular}[b]{l|l|c|r|cc}
    \hline
    \multirow{2}{*}{Method} &\multirow{2}{*}{Network} &  \multirow{2}{*}{Pre-training} &\multirow{2}{*}{Input Size} & \multicolumn{2}{c}{Accuracy (\%)} \\
     &&&& Top-1 & Top-5 \\ \hline
    C3D~\cite{Hara_2018_CVPR} &ResNet-101  &None &16$\times$112$\times$112 &62.8 &83.9 \\
    C3D~\cite{Hara_2018_CVPR} &ResNeXt-101 &None &16$\times$112$\times$112 &65.1 &85.7 \\
    ARTNet~\cite{Wang_2018_CVPR} &ResNet-18 &None &16$\times$112$\times$112 &69.2 &88.3 \\
    STC~\cite{diba2018spatio-temporal} &ResNeXt-101 &None  &32$\times$112$\times$112 &68.7 &88.5 \\
    I3D~\cite{kinetics} &Inception  &ImageNet &64$\times$224$\times$224  & 71.1$^*$ & 89.3$^*$ \\
    R(2+1)D~\cite{Tran2017A} &Custom &None &8$\times$112$\times$112  &72.0 &90.0 \\
    R(2+1)D~\cite{Tran2017A} &Custom &Sports-1M &8$\times$112$\times$112  &74.3 &91.4 \\ 
    S3D-G~\cite{Xie2018Rethinking} &Inception &ImageNet &64$\times$224$\times$224 &74.7 &\textbf{93.4} \\
    NL I3D~\cite{Wang2017Non} &ResNet-101  &ImageNet &32$\times$224$\times$224 &76.0 &92.1 \\
    NL I3D~\cite{Wang2017Non} &ResNet-101  &ImageNet &128$\times$224$\times$224 &\textbf{77.7} &93.3 \\ \hline
    CoST &ResNet-101 &ImageNet &8$\times$224$\times$224 &75.5 &92.0 \\
    CoST &ResNet-101 &ImageNet &32$\times$224$\times$224 & \textbf{77.5} & \textbf{93.2} \\ \hline
  \end{tabular}
  \end{center}
  \caption{Comparison with the state-of-the-arts on the validation set of Kinetics. For fair comparison, only results based on the RGB modality are listed. All the numbers are \emph{single-model} results. $^*$ indicates results on the test set.}
  \label{table:kintics_stateoftheart}
\end{table*}

\begin{table}[t]
  \begin{center}
  \begin{tabular}[b]{l|cc}
    \hline
    \multirow{2}{*}{Method}  & \multicolumn{2}{c}{Accuracy (\%)} \\
     & Top-1 & Top-5 \\ \hline
     ResNet-50-Scratch~\cite{monfortmoments} &23.7 &46.7\\
     ResNet-50-ImageNet~\cite{monfortmoments} &27.2 &51.7\\
     \color{gray} SoundNet-Audio~\cite{monfortmoments} &\color{gray} 7.6 &\color{gray} 18.0\\
     \color{gray} TSN-Flow~\cite{monfortmoments}  &\color{gray} 15.7 &\color{gray} 34.7\\
     \color{gray} RGB+Flow+Audio~\cite{monfortmoments} &\color{gray} 30.4 &\color{gray} 55.9\\\hline
     CoST (ResNet-50, 8 frames)  &30.1 &57.2 \\
     CoST (ResNet-101, 8 frames) &31.5 &57.9 \\
     CoST (ResNet-101, 32 frames) &\textbf{32.4} &\textbf{60.0} \\ \hline
  \end{tabular}
  \end{center}
  \caption{Comparison with the state-of-the-arts on the validation set of Moments in Time. Methods marked in gray exploit additional modalities, \eg audio and optical flow.}
  \label{table:moments_stateoftheart}
\end{table}

Besides the 8-frame model, we also train a model with a higher temporal resolution, i.e. 32 frames. On Moments in time, the 32 input frames are sampled from 64 continuous frames mentioned earlier. While on Kinetics, we sample 32 frames from a clip of 128 frames considering that videos in this dataset is longer than those in Moments in Time. The 32-frame model is fine-tuned from the 8-frame model, where the parameters of BN layers~\cite{ioffe2015batch} are frozen.

On the Moments in Time dataset, Table~\ref{table:moments_stateoftheart} shows a comparison of the proposed CoST with existing methods. CoST improves the ResNet-50 C2D baseline reported in~\cite{monfortmoments} by 2.9\% and 5.5\% in terms of top-1 and top-5 accuracies respectively. While ResNet-101 based CoST with 32 input frames achieves 32.4\% top-1 accuracy and 60.0\% top-5 accuracy. Notably, based on the RGB modality only, our model outperforms the ensemble result of multiple modalities (i.e. RGB, optical flow and audio) in~\cite{monfortmoments} by a large margin. With an ensemble of multiple models and modalities, we achieve 52.91\% average accuracy on the test set, which won the 1st place in the Moments in Time Challenge 2018.

On the Kinetics dataset, CoST achieves state-of-the-art performance. As shown in Table~\ref{table:kintics_stateoftheart}, CoST has a clear advantage over C3D~\cite{Hara_2018_CVPR} and its variants, \eg I3D~\cite{kinetics}, R(2+1)D~\cite{Tran2017A} and S3D-G~\cite{Xie2018Rethinking}. Compared with NL I3D~\cite{Wang2017Non}, which is a strong baseline, CoST is also superior at various temporal resolutions.

\subsection{Importance of Different Views}
\begin{figure*}[t]
  \begin{center}
  \includegraphics[width=0.45\textwidth]{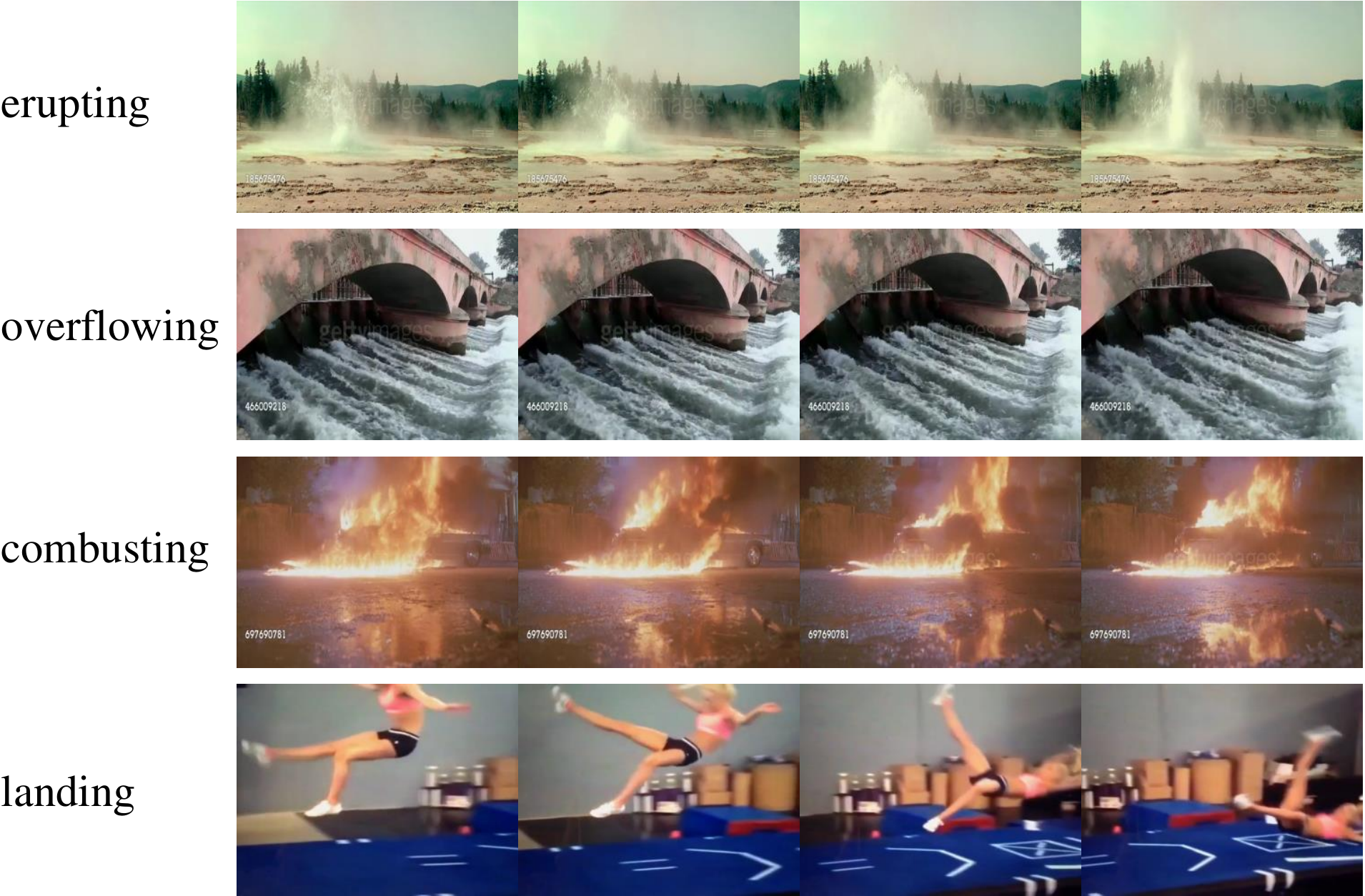}~~~~
  \includegraphics[width=0.45\textwidth]{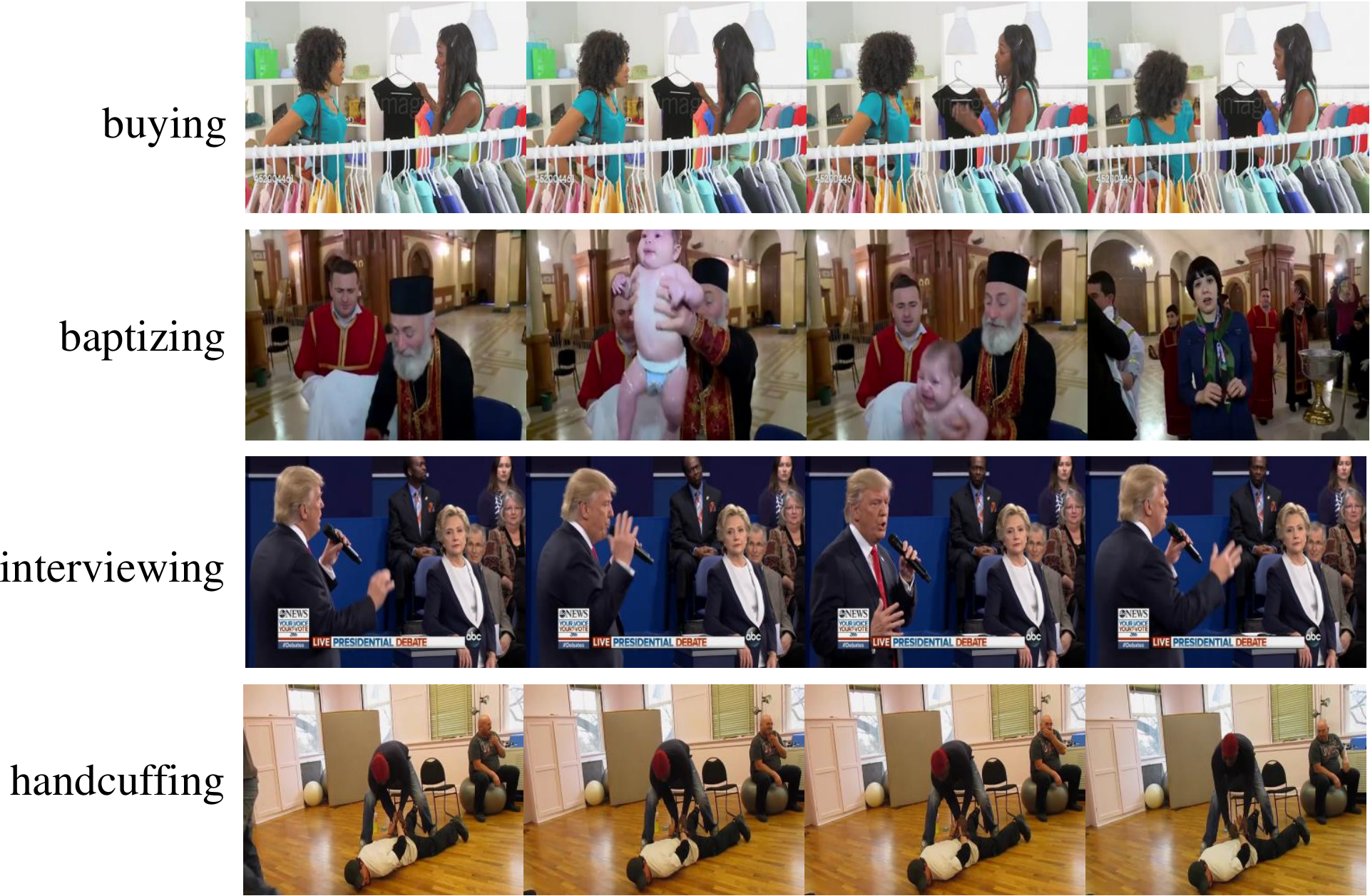}
  \end{center}
  \caption{Left: actions for which temporal feature matters. Right: actions for which temporal feature is less important.
  \label{fig:examples}}
\end{figure*}

\begin{figure}[t]
  \begin{center}
  \includegraphics[width=\linewidth]{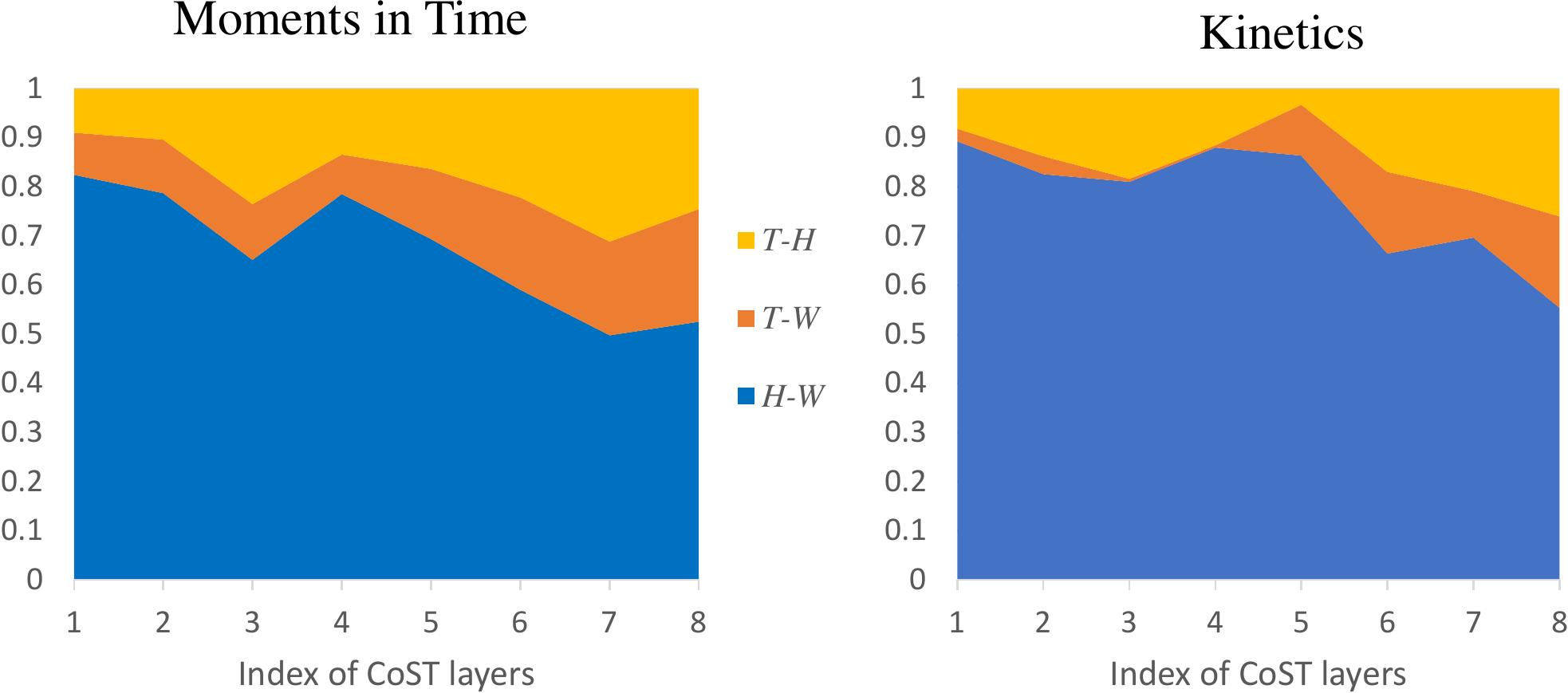}
  \end{center}
  \caption{Distribution of the mean coefficient among the three views in CoST layers of various depths.
  \label{fig:moments_kinetics}}
\end{figure}

By investigating the magnitude of the learned coefficients, we are able to quantify the contribution of different views. Specifically, for each CoST layer, the mean coefficient of each view is computed on the validation set. The mean coefficient of the $H\mhyphen W$ view measures the importance of appearance feature, while those of the $T\mhyphen W$ and $T\mhyphen H$ views measure the importance of temporal motion cues.

The overall importance of each view can be measured by averaging the mean coefficients of all CoST layers. On Moments in Time, the mean coefficients of the $H\mhyphen W$, $T\mhyphen W$ and $T\mhyphen H$ views are 0.67, 0.14 and 0.19 respectively. While on Kinetics they are 0.77, 0.08 and 0.15. Hence, spatial feature plays a major role on both datasets. And the Moments in Time dataset depends more on temporal feature to discriminate different actions than Kinetics.

Figure~\ref{fig:moments_kinetics} shows the coefficient distribution among the three views in all CoST layers of the ResNet-50 based CoST. From shallow layer to deep layer, a clear trend is observed on both datasets. That is, the contribution of spatial feature declines, while that of temporal feature rises. In other words, the closer to top of a network, the more important the temporal feature is, suggesting that the model tends to learn temporal feature based on high-level spatial feature. This also verifies the conclusion in~\cite{Xie2018Rethinking} that temporal representation learning on high-level semantic features is more useful than low-level features.

Furthermore, we analyze the importance of spatial and temporal features for each action category on the Moments in Time dataset. We sum up the mean coefficients of temporal related views and sort all categories by it. As shown in Figure~\ref{fig:examples}, for actions such as erupting, storming, overflowing, combusting and landing, temporal motion information is very important. On the contrary, for actions such as baptizing, handcuffing / arresting, interviewing, buying and paying, temporal feature is less important. These actions can either be easily recognized by appearance, or the temporal evolutions are not very helpful for classification. For example, for buying and interviewing various motion patterns exist within the same category and they may be easily confused between different actions, which makes the motion cues not discriminative.

In summary, with the proposed CoST, we are able to quantitatively analyze the importance of spatial and temporal features. In particular, we observe that the bottom layers of the network focus more on spatial feature learning, while the top layers attend more to temporal feature aggregation. Besides, some actions are easier to recognize based on the underlying objects and their interactions (\eg geometric relation) rather than motion cues. This indicates that the current spatiotemporal feature learning approaches may not be optimal, and we expect more efforts on this problem.

\section{Discussion}
For video analysis, how to encode spatiotemporal features effectively and efficiently is still an open question. In this work, we propose to use weight-shared 2D convolutions for simultaneous spatial and temporal feature encoding. Although we empirically verify that weight sharing brings performance gain, one big question behind is whether the temporal dimension $T$ can be cast as a normal spatial dimension (like depth) or not. Intuitively, spatial appearance feature and temporal motion cue belong to two different modalities of information. What motivates us to learn them collaboratively is the visualization of different views as shown in Figure~\ref{fig:conception}. Interestingly, our positive results indicate that at least to some extent, they share similar characteristics and can be jointly learned using a single network with identical network architecture and shared convolution kernels. In physics, according to Minkowski spacetime~\cite{minkowski1908space}, the three-dimensional space and one-dimensional time can be unified as a four-dimensional continuum. Our finding might be explained and supported by the spacetime model in the context of feature representation learning.

\section{Conclusion}
Feature learning from 3D volumetric data is the major challenge for action recognition in videos. In this paper, we propose a novel feature learning operation, which learns spatiotemporal features collaboratively from multiple views. It can be easily used as a drop-in replacement for C2D and C3D. Experiments on large-scale benchmarks validate the superiority of the proposed architecture over existing methods. Based on the learned coefficients of different views, we are able to take a peek at the individual contribution of spatial and temporal features for classification. A systematic analysis indicates some promising directions on the design of algorithm, which we will leave as future work.

{\small
\bibliographystyle{ieee}
\bibliography{main}

\begin{thebibliography}{10}\itemsep=-1pt

\bibitem{ST-SIFT}
M.~Al~Ghamdi, L.~Zhang, and Y.~Gotoh.
\newblock Spatio-temporal sift and its application to human action
  classification.
\newblock In {\em Computer Vision -- ECCV 2012. Workshops and Demonstrations},
  2012.

\bibitem{kinetics}
J.~Carreira and A.~Zisserman.
\newblock Quo vadis, action recognition? a new model and the kinetics dataset.
\newblock In {\em The IEEE Conference on Computer Vision and Pattern
  Recognition (CVPR)}, July 2017.

\bibitem{diba2018spatio-temporal}
A.~Diba, M.~Fayyaz, V.~Sharma, M.~M. Arzani, R.~Yousefzadeh, J.~Gall, and
  L.~Van~Gool.
\newblock Spatio-temporal channel correlation networks for action
  classification.
\newblock {\em ECCV}, pages 299--315, 2018.

\bibitem{donahue2017long-term}
J.~Donahue, L.~A. Hendricks, M.~Rohrbach, S.~Venugopalan, S.~Guadarrama,
  K.~Saenko, and T.~Darrell.
\newblock Long-term recurrent convolutional networks for visual recognition and
  description.
\newblock {\em TPAMI}, 39(4):677--691, 2017.

\bibitem{feichtenhofer2016spatiotemporal}
C.~Feichtenhofer, A.~Pinz, and R.~P. Wildes.
\newblock Spatiotemporal residual networks for video action recognition.
\newblock {\em neural information processing systems}, pages 3468--3476, 2016.

\bibitem{Feichtenhofer2017Spatiotemporal}
C.~Feichtenhofer, A.~Pinz, and R.~P. Wildes.
\newblock Spatiotemporal multiplier networks for video action recognition.
\newblock In {\em CVPR}, pages 7445--7454, 2017.

\bibitem{Hara_2018_CVPR}
K.~Hara, H.~Kataoka, and Y.~Satoh.
\newblock Can spatiotemporal 3d cnns retrace the history of 2d cnns and
  imagenet?
\newblock In {\em CVPR}, June 2018.

\bibitem{He2015Deep}
K.~He, X.~Zhang, S.~Ren, and J.~Sun.
\newblock Deep residual learning for image recognition.
\newblock {\em computer vision and pattern recognition}, pages 770--778, 2016.

\bibitem{He_2017_ICCV}
Y.~He, X.~Zhang, and J.~Sun.
\newblock Channel pruning for accelerating very deep neural networks.
\newblock In {\em ICCV}, Oct 2017.

\bibitem{ioffe2015batch}
S.~Ioffe and C.~Szegedy.
\newblock Batch normalization: Accelerating deep network training by reducing
  internal covariate shift.
\newblock In {\em International Conference on Machine Learning}, pages
  448--456, 2015.

\bibitem{sports-1M}
A.~Karpathy, G.~Toderici, S.~Shetty, T.~Leung, R.~Sukthankar, and F.~F. Li.
\newblock Large-scale video classification with convolutional neural networks.
\newblock In {\em Computer Vision and Pattern Recognition}, pages 1725--1732,
  2014.

\bibitem{Klaser2010A}
A.~Klaser.
\newblock A spatiotemporal descriptor based on 3d-gradients.
\newblock In {\em British Machine Vision Conference, September}, 2010.

\bibitem{krizhevsky2012imagenet}
A.~Krizhevsky, I.~Sutskever, and G.~E. Hinton.
\newblock Imagenet classification with deep convolutional neural networks.
\newblock In {\em NIPS}, 2012.

\bibitem{Laptev2005On}
I.~Laptev.
\newblock On space-time interest points. ijcv.
\newblock {\em International Journal of Computer Vision}, 64(2):107--123, 2005.

\bibitem{li2017pruning}
H.~Li, A.~Kadav, I.~Durdanovic, H.~Samet, and H.~P. Graf.
\newblock Pruning filters for efficient convnets.
\newblock {\em International Conference on Learning Representations}, 2017.

\bibitem{minkowski1908space}
H.~Minkowski et~al.
\newblock Space and time.
\newblock {\em The principle of relativity}, pages 73--91, 1908.

\bibitem{monfortmoments}
M.~Monfort, B.~Zhou, S.~A. Bargal, T.~Yan, A.~Andonian, K.~Ramakrishnan,
  L.~Brown, Q.~Fan, D.~Gutfruend, C.~Vondrick, et~al.
\newblock Moments in time dataset: one million videos for event understanding.

\bibitem{ng2015beyond}
J.~Y. Ng, M.~J. Hausknecht, S.~Vijayanarasimhan, O.~Vinyals, R.~Monga, and
  G.~Toderici.
\newblock Beyond short snippets: Deep networks for video classification.
\newblock {\em computer vision and pattern recognition}, pages 4694--4702,
  2015.

\bibitem{Qiu2017Learning}
Z.~Qiu, T.~Yao, and T.~Mei.
\newblock Learning spatio-temporal representation with pseudo-3d residual
  networks.
\newblock In {\em ICCV}, Oct 2017.

\bibitem{russakovsky2015imagenet}
O.~Russakovsky, J.~Deng, H.~Su, J.~Krause, S.~Satheesh, S.~Ma, Z.~Huang,
  A.~Karpathy, A.~Khosla, M.~Bernstein, et~al.
\newblock Imagenet large scale visual recognition challenge.
\newblock {\em International Journal of Computer Vision}, 115(3):211--252,
  2015.

\bibitem{Scovanner2007A}
P.~Scovanner, S.~Ali, and M.~Shah.
\newblock A 3-dimensional sift descriptor and its application to action
  recognition.
\newblock In {\em International Conference on Multimedia}, pages 357--360,
  2007.

\bibitem{Shao_2016_CVPR}
J.~Shao, C.-C. Loy, K.~Kang, and X.~Wang.
\newblock Slicing convolutional neural network for crowd video understanding.
\newblock In {\em CVPR}, June 2016.

\bibitem{simonyan2014two-stream}
K.~Simonyan and A.~Zisserman.
\newblock Two-stream convolutional networks for action recognition in videos.
\newblock {\em neural information processing systems}, pages 568--576, 2014.

\bibitem{Du2014Learning}
D.~Tran, L.~D. Bourdev, R.~Fergus, L.~Torresani, and M.~Paluri.
\newblock Learning spatiotemporal features with 3d convolutional networks.
\newblock {\em international conference on computer vision}, pages 4489--4497,
  2015.

\bibitem{Tran2017A}
D.~Tran, H.~Wang, L.~Torresani, J.~Ray, Y.~LeCun, and M.~Paluri.
\newblock A closer look at spatiotemporal convolutions for action recognition.
\newblock In {\em CVPR}, June 2018.

\bibitem{vaswani2017attention}
A.~Vaswani, N.~Shazeer, N.~Parmar, J.~Uszkoreit, L.~Jones, A.~N. Gomez,
  L.~Kaiser, and I.~Polosukhin.
\newblock Attention is all you need.
\newblock {\em NIPS}, pages 5998--6008, 2017.

\bibitem{Wang2013Dense}
H.~Wang, A.~Kl{\"a}ser, C.~Schmid, and C.~L. Liu.
\newblock Dense trajectories and motion boundary descriptors for action
  recognition.
\newblock {\em IJCV}, 103(1):60--79, 2013.

\bibitem{Wang2014Action}
H.~Wang and C.~Schmid.
\newblock Action recognition with improved trajectories.
\newblock In {\em IEEE International Conference on Computer Vision}, pages
  3551--3558, 2014.

\bibitem{Wang_2018_CVPR}
L.~Wang, W.~Li, W.~Li, and L.~Van~Gool.
\newblock Appearance-and-relation networks for video classification.
\newblock In {\em CVPR}, June 2018.

\bibitem{Wang2017Non}
X.~Wang, R.~Girshick, A.~Gupta, and K.~He.
\newblock Non-local neural networks.
\newblock In {\em CVPR}, June 2018.

\bibitem{xie2017all}
D.~Xie, J.~Xiong, and S.~Pu.
\newblock All you need is beyond a good init: Exploring better solution for
  training extremely deep convolutional neural networks with orthonormality and
  modulation.
\newblock In {\em CVPR}, pages 6176--6185, 2017.

\bibitem{Xie2018Rethinking}
S.~Xie, C.~Sun, J.~Huang, Z.~Tu, and K.~Murphy.
\newblock Rethinking spatiotemporal feature learning: Speed-accuracy trade-offs
  in video classification.
\newblock In {\em ECCV}, 2018.

\bibitem{zhao2017temporal}
Y.~Zhao, Y.~Xiong, L.~Wang, Z.~Wu, X.~Tang, and D.~Lin.
\newblock Temporal action detection with structured segment networks.
\newblock {\em international conference on computer vision}, pages 2933--2942,
  2017.

\bibitem{zhu2017hidden}
Y.~Zhu, Z.~Lan, S.~D. Newsam, and A.~G. Hauptmann.
\newblock Hidden two-stream convolutional networks for action recognition.
\newblock {\em arXiv: Computer Vision and Pattern Recognition}, 2017.

\end{thebibliography}
}

\end{document}